\documentclass[10pt,conference]{IEEEtran}
\IEEEoverridecommandlockouts
\usepackage{cite}
\usepackage{amsmath,amssymb,amsfonts}
\usepackage{booktabs}
\usepackage{algorithmic}
\usepackage{graphicx}
\usepackage{textcomp}
\usepackage{bm}
\usepackage{hyperref}
\usepackage{xcolor}
\def\BibTeX{{\rm B\kern-.05em{\sc i\kern-.025em b}\kern-.08em
    T\kern-.1667em\lower.7ex\hbox{E}\kern-.125emX}}
\begin{document}

\title{An Analysis of Regularization and Fokker--Planck Residuals in Diffusion Models for Image Generation\\

\thanks{The authors acknowledge financial support from project PID2022-139856NB-I00 funded by MCIN/ AEI / 10.13039/501100011033 /
FEDER, UE and project, IDEA-CM (TEC-2024/COM-89) from the Autonomous
Community of Madrid and from the ELLIS Unit Madrid.  The authors
acknowledge computational support from the Centro de Computación
Científica-Universidad Autónoma de Madrid (CCC-UAM).}
}

\author{\IEEEauthorblockN{Onno Niemann}
\IEEEauthorblockA{\textit{Universidad Autónoma de Madrid}\\
onno.niemann@uam.es}
\and
\IEEEauthorblockN{Gonzalo Martínez Muñoz}
\IEEEauthorblockA{\textit{Universidad Autónoma de Madrid}\\
gonzalo.martinez@uam.es}
\and
\IEEEauthorblockN{Alberto Suárez Gonzalez}
\IEEEauthorblockA{\textit{Universidad Autónoma de Madrid}\\
alberto.suarez@uam.es}
}

\maketitle

\begin{abstract}
Recent work has shown that diffusion models trained with the denoising score matching (DSM) objective often violate the Fokker--Planck (FP) equation that governs the evolution of the true data density. 
Directly penalizing these deviations in the objective function reduces their magnitude but introduces a significant computational overhead.
It is also observed that enforcing strict adherence to the FP equation does not necessarily lead to improvements in the quality of the generated samples, as often the best results are obtained with weaker FP regularization.
In this paper, we investigate whether simpler penalty terms can provide similar benefits. 
We empirically analyze several lightweight regularizers, study their effect on FP residuals and generation quality, and show that the benefits of FP regularization are available at substantially lower computational cost. Our code is available at \url{https://github.com/OnnoNiemann/fp_diffusion_analysis}.
\end{abstract}

\begin{IEEEkeywords}
Diffusion Models, Score-Based Generative Models, Regularization, Fokker--Planck Equation
\end{IEEEkeywords}

\section{Introduction}

Diffusion models (DMs) have emerged as a powerful class of generative models, achieving state-of-the-art performance in image synthesis and likelihood-based evaluation \cite{song2021sde}.
Initially, DMs were formulated in discrete time, with noise added and removed over a finite sequence of steps. Two closely related frameworks shaped these developments: Score Matching with Langevin Dynamics (SMLD) \cite{song2019generative} and Denoising Diffusion Probabilistic Models (DDPMs) \cite{ho2020ddpm}. SMLD leverages annealed Langevin dynamics, a process that generates samples by moving through a sequence of discrete noise levels, while following the score. DDPM uses a framework describing noise injection as a Markov chain and trains a neural network to remove the noise. 

Both of these discrete-time models were later unified under a continuous-time SDE framework \cite{song2021sde}, in which the forward noising process that progressively adds noise to clean images, and the reverse, denoising process, are both described by Stochastic Differential Equations (SDEs) \cite{anderson1982diff}. The forward process maps clean images to a simple, known prior distribution and the denoising process enables the generation of new images by starting from pure random noise and iteratively denoising until the generated samples converge to the distribution of the training data. 
This iterative noise removal requires access to the score of the data distribution, which is typically learned by training a neural network (called a noise-conditional score model) by Denoising Score Matching (DSM). Minimizing the DSM objective yields an accurate approximation of the scores of data distributions at different noise levels and enables sampling via iteratively following the reverse SDE \cite{song2021sde}.

Despite their success, recent work has uncovered a fundamental limitation of score-based generative models. The forward noising process follows a diffusion SDE and therefore the evolution of the ground truth marginal densities over noise levels is described by the Fokker--Planck (FP) equation \cite{Oksendal2003}. However, the learned score approximations obtained via DSM often violate the corresponding condition that the FP equation imposes on the score \cite{Lai2023FPDiffusion, plainer2025consistent}. Explicitly enforcing this equation by including a FP-based regularization term in the training objective reduces the violation. However, empirical results show that strong FP regularization does not always yield the highest-quality samples. 
\cite{Lai2023FPDiffusion}.

Moreover, computing the second-order derivatives that appear in the FP penalty is computationally demanding, and can significantly increase the training time with respect to non-regularized diffusion models \cite{Lai2023FPDiffusion}. 
On top of computational challenges, the FP penalty consists of many components and their interactions, making it difficult to attribute improvements to individual terms. This raises the question, whether the observed benefits of FP regularization require the full complexity of enforcing the FP equation, or can be achieved by leveraging more general regularization effects at lower computational expense and with more interpretable penalties.

The goal of this paper is to systematically investigate a set of such penalty terms. 
We evaluate their impact on FP residuals, score behavior across noise levels, and downstream sample quality.
The results of this study show empirically that simpler regularizers can deliver benefits comparable to FP regularization at a substantially lower computational cost.

\section{Background}
\subsection{Score-based diffusion models}

The goal of generative models is to capture the structure, variability, and dependencies in the training data, making it possible to produce realistic and diverse samples that resemble the original ones.
In image generation, the data available for induction are images whose distribution, $p_0(\bm{x})$, is unknown.
Score-based generative models consist of a forward noise-injection and a backward denoising process. 
In the forward process, noise is injected into the original images according to a forward SDE
\begin{equation}\label{FORW}
    d\bm{x}=\bm{f}(\bm{x},t)dt+g(t)d\bm{w}_t ,
\end{equation}
where $\bm{x}\in \mathbb{R}^D$ denotes an image with $D$ pixels,  $\bm{f}(\bm{x}, t) \in  \mathbb{R}^D$ is the deterministic drift term, $g(t) \in \mathbb{R}$ is a scalar diffusion term that controls the strength of the noise injected at $t$, and \(\bm{w}_t\) denotes a Wiener process (standard Brownian motion). 

The forward process progressively injects noise, according to \eqref{FORW}, starting from \(\bm{x}(0) \sim p_0(\bm{x}) \), one of the images of the training dataset. 
This yields a sequence of images $\left\{\bm{x}(t), t \in [0,T] \right\}$ where larger values of \(t\) correspond to higher noise levels. 
This sequence eventually converges to $\bm{x}(T)\sim p_T(\bm{x})$, a simple, known reference distribution. 
In practice, $\bm{f}(\bm{x},t)$ and $g(t)$ are chosen so that both the reference distribution, $p_T(\bm{x})$, 
and the conditional distribution $p_t(\bm{x}(t)\mid \bm{x}(0))$ are Gaussian.
If the forward SDE has a Gaussian transition kernel,
\begin{equation}\label{conditional}
    p_t(\bm{x}(t)\mid \bm{x}(0)) = \mathcal{N}\!\left(\bm{\mu}(\bm{x}(0), t),\, \sigma^2(t) \bm{I}\right),
\end{equation}
where \(\bm{\mu}(\bm{x}(0), t)\) and \(\sigma(t)\) are determined by the choice of the functions $\bm{f}(\bm{x},t)$ and $g(t)$,
the corrupted image at time (noise level) \(t\) can be obtained in one step
\begin{equation}\label{eq:onestep}
    \bm{x}(t)=\bm{\mu}(\bm{x}(0), t)+\sigma(t)\bm{z}, \quad z \sim \mathcal{N}(0, I).
\end{equation}
Common choices of $\bm{f}(\bm{x},t)$ and $g(t)$ are the variance-preserving (VP) and variance-exploding (VE) SDEs \cite{song2021sde}. 

To generate new samples from $p_0(\bm x)$, the distribution of the original images, the noise injection process can be reversed by solving the backward SDE \cite{anderson1982diff}
\begin{equation}\label{BACKW}
    d\bm{x}=[\bm{f}(\bm{x},t)-g(t)^2 \nabla_{\bm{x}} \log p_t(\bm{x})]dt+g(t)d\bm{w}_t,
\end{equation}
where $\bm{w}_t$ is standard Brownian motion going backward in time and $\nabla_{\bm{x}} \log p_t(\bm{x})$ is the Stein score function, i.e. the gradient of the log-probability of the data distribution at time $t$. 
The generating process starts from an image sampled from the reference distribution at time $T$, $\overleftarrow{\bm{x}}(T)\sim p_T(\bm{x})$.
From this sample the noise is iteratively removed by integrating the backwards SDE from \(T\) until \(0\), so that $\overleftarrow{\bm{x}}(0)\sim p_0(\bm{x})$.
This implies that the generated image exhibits a statistical resemblance to the original ones in the training data. 

\begin{figure}[bt]
    \centering
    \includegraphics[width=\linewidth]{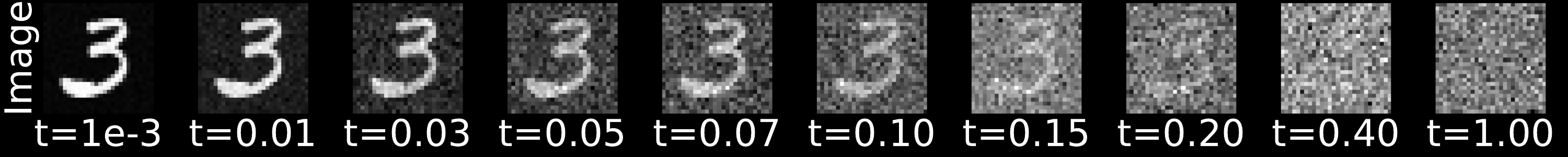}\\
\hspace{1mm}
    \includegraphics[width=\linewidth]{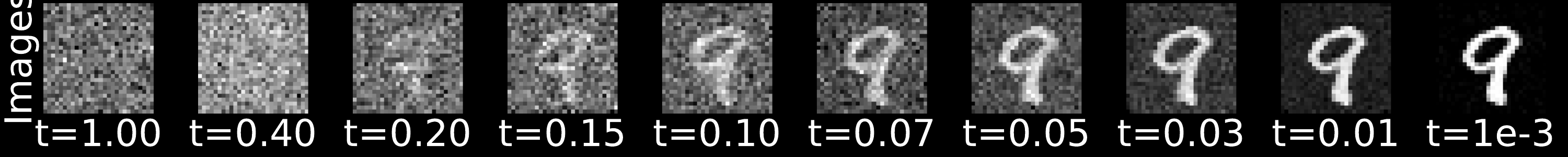}
    \caption{Noise injection in the forward process following \eqref{FORW} (top) and image generation from noise according to \eqref{BACKW} (bottom).}
    \label{fig:for_back}
\end{figure}

In order to perform the backward process, it is necessary to estimate the term $\nabla_{\bm{x}} \log p_t(\bm{x})$. However, learning the score function directly is not possible, since the marginal $p_t(\bm{x})$ and the score are unknown. 
Notwithstanding, DSM provides an estimator for this score that leverages the known conditional given in \eqref{conditional}. 
This allows us to approximate the score using a neural score network, $\bm{s}(\bm{x}(t), t; \bm{\theta})$.
Examples of the backward and forward processes are illustrated in Fig.~\ref{fig:for_back}. 

\subsection{Denoising Score Matching}
The goal of score matching is to train a model $s(\bm{x}(t),t; \bm{\theta})$ to approximate $\nabla_{\bm{x}} \log p_t(\bm{x})$ without requiring access to the normalization constant of $p(\bm{x})$. 
Conceptually, this is achieved by training a neural network with entries and outputs in \(\mathbb{R}^D\) to map noisy images, $\bm{x}(t)$, to the correct scores, $\nabla_{\bm{x}} \log p_t(\bm{x})$. 
However, as the ground truth score function is unknown in the context of images, this explicit score matching idea is replaced by Denoising Score Matching, which learns the score of noisy images conditioned on clean ones
\begin{equation}
    \begin{aligned}
        L_{\text{DSM}}(\bm{\theta})
        &= \mathbb{E}_{\bm{x}(0),t,\bm{x}(t)\mid \bm{x}(0)}
        \Big[\big\| \bm{s}(\bm{x}(t), t; \bm{\theta}) \\&- \nabla_{\bm{x}(t)} \log p_t(\bm{x}(t)\mid \bm{x}(0)) \big\|_2^2\Big]
    \label{eq:lossdsm}.
    \end{aligned}
\end{equation}
This function can be used in practice, since we know the score of the conditional distribution \eqref{conditional}
\begin{equation}
    \nabla_{\bm{x}(t)} \log p_t(\bm{x}(t)\mid \bm{x}(0))
= -\frac{\bm{x}(t) - \bm{\mu}(\bm{x}(0), t)}{\sigma(t)^{2}}=-\frac{\bm{z}}{\sigma(t)},
\end{equation}
where \eqref{eq:onestep} is applied in the last step. Substituting this result into \eqref{eq:lossdsm} and multiplying by $\sigma(t)$
for improved numerical stability gives the DSM loss used to train the score network
\begin{equation} \label{LDSM}
    L_{\text{DSM}}(\bm{\theta}) = \mathbb{E}_{\bm{x}(0),t,\bm{z}} \left[ \left\| \sigma(t) \, \bm{s}(\bm{x}(t), t; \bm{\theta}) + \bm{z} \right\|_2^2 \right].
\end{equation}

\subsection{Fokker--Planck and Score Fokker--Planck Equation}
The forward noising process defined in \eqref{FORW} is a diffusion process in which the evolution of the (unknown) ground truth data density over noise levels follows the FP equation \cite{Oksendal2003}
\begin{equation}\label{FP}
    \begin{aligned}
        &\partial_t p_t(\bm{x}) = -\sum_{j=1}^D \partial_{x_j}\big(\bm{F}_j(\bm{x},t)\,p_t(\bm{x})\big),
    \end{aligned}
\end{equation}
where $\bm{F}(\bm{x},t) = \bm{f}(\bm{x},t) - \tfrac{1}{2} g^2(t)\,\nabla_{\bm{x}} \log p_t(\bm{x})$.
However, calculating this FP equation is not feasible in the context of score-based generative models, since they do not model $p_t(\bm{x})$. For this reason, a corresponding condition on the score is derived in \cite{Lai2023FPDiffusion}, the score FP equation 
\begin{equation}\label{SFP}
    \begin{aligned}
        \partial_t \bm{s}(\bm{x},t) &= \nabla_{\bm{x}}\Bigg[
        \tfrac{1}{2}g^2(t)\,\mathrm{div}_{\bm{x}} \bm{s}(\bm{x},t)
        + \tfrac{1}{2}g^2(t)\,\|\bm{s}(\bm{x},t)\|_2^2\\
        & \qquad \
        - \langle \bm{f}(\bm{x},t),\,\bm{s}(\bm{x},t)\rangle
        - \mathrm{div}_{\bm{x}} \bm{f}(\bm{x},t)\Bigg].
    \end{aligned}
\end{equation}
By defining 
\begin{equation}\label{eq:L_helper}
    \mathcal{L}[\cdot]
    := \tfrac{1}{2} g^2\, \mathrm{div}_{\bm{x}}(\cdot)
    + \tfrac{1}{2} g^2\, \lVert \cdot \rVert_2^2
    - \langle \bm{f}, \cdot \rangle
    - \mathrm{div}_{\bm{x}}\!\bigl(\bm{f}\bigr),
\end{equation} 
\eqref{SFP} can be written as $\partial_t \bm{s}(\bm{x},t) = \nabla_{\bm{x}} \mathcal{L}[\bm{s}](\bm{x},t)$.

An interesting finding of \cite{Lai2023FPDiffusion} is that while \eqref{SFP} must hold for the temporal evolution of the unknown ground truth score, score models trained using the DSM loss \eqref{LDSM} often violate this condition. The deviation in the evolution of the approximate scores from the ones given by the FP equation can be quantified by subtracting the left and right side terms of \eqref{SFP} 
\begin{equation}\label{FPErr}
    \bm{\varepsilon}[\bm{s}](\bm{x},t; \bm{\theta})
    := \partial_t \bm{s}(\bm{x},t; \bm{\theta}) - \nabla_{\bm{x}} \mathcal{L}[\bm{s}](\bm{x},t; \bm{\theta}).
\end{equation}
The average of the squared L2 norm of this error over the pixel dimension gives a time-dependent FP residual that measures the deviation from the score FP equation
\begin{equation}\label{RFP}
    r_{\mathrm{FP}}[\bm{s}](t; \bm{\theta})
    := \frac{1}{D}\,\mathbb{E}_{\bm{x}(0),\bm{x}(t)\mid \bm{x}(0)}
    \big[\|\bm{\varepsilon}[\bm{s}](\bm{x},t;\bm{\theta})\|_{2}^{2}\big].
\end{equation}
This residual is found to be large, especially for small noise levels \cite{Lai2023FPDiffusion}.

\section{Regularization and Penalty Terms}
The observation that diffusion models trained with DSM often violate the FP equation, suggests that adding a regularization term to the loss may produce models that adhere more closely to the FP equation, which should yield better score estimates. 
This section introduces some existing and proposed regularization/penalization terms.

\subsection{Fokker--Planck Penalty}
In \cite{Lai2023FPDiffusion}, a regularization term based on the error given in \eqref{FPErr} is proposed
\begin{equation}\label{PFP}
    P_{\mathrm{FP}}:= \frac{1}{D}\,\mathbb{E}_{\bm{x}(0),t,\bm{z}} \big[\|\bm{\varepsilon}[\bm{s}](\bm{x},t;\bm{\theta})\|_{2}\big].
\end{equation}
This norm involves the computation of many terms that are both complicated to interpret and to compute. Specifically, it involves the computation of the squared norm of the gradients of all terms in \eqref{eq:L_helper}, plus their interactions. 

While this penalty is theoretically well-motivated, it is not obvious which of its components are responsible for improved empirical performance. 
This motivates our investigation of simpler penalties that isolate individual mechanisms, such as controlling score magnitude or smoothness, without explicitly enforcing the full score FP equation.

\subsection{Deeper Understanding of Score and Divergence}
\begin{figure}[tb]
    \centering
    \includegraphics[width=\linewidth]{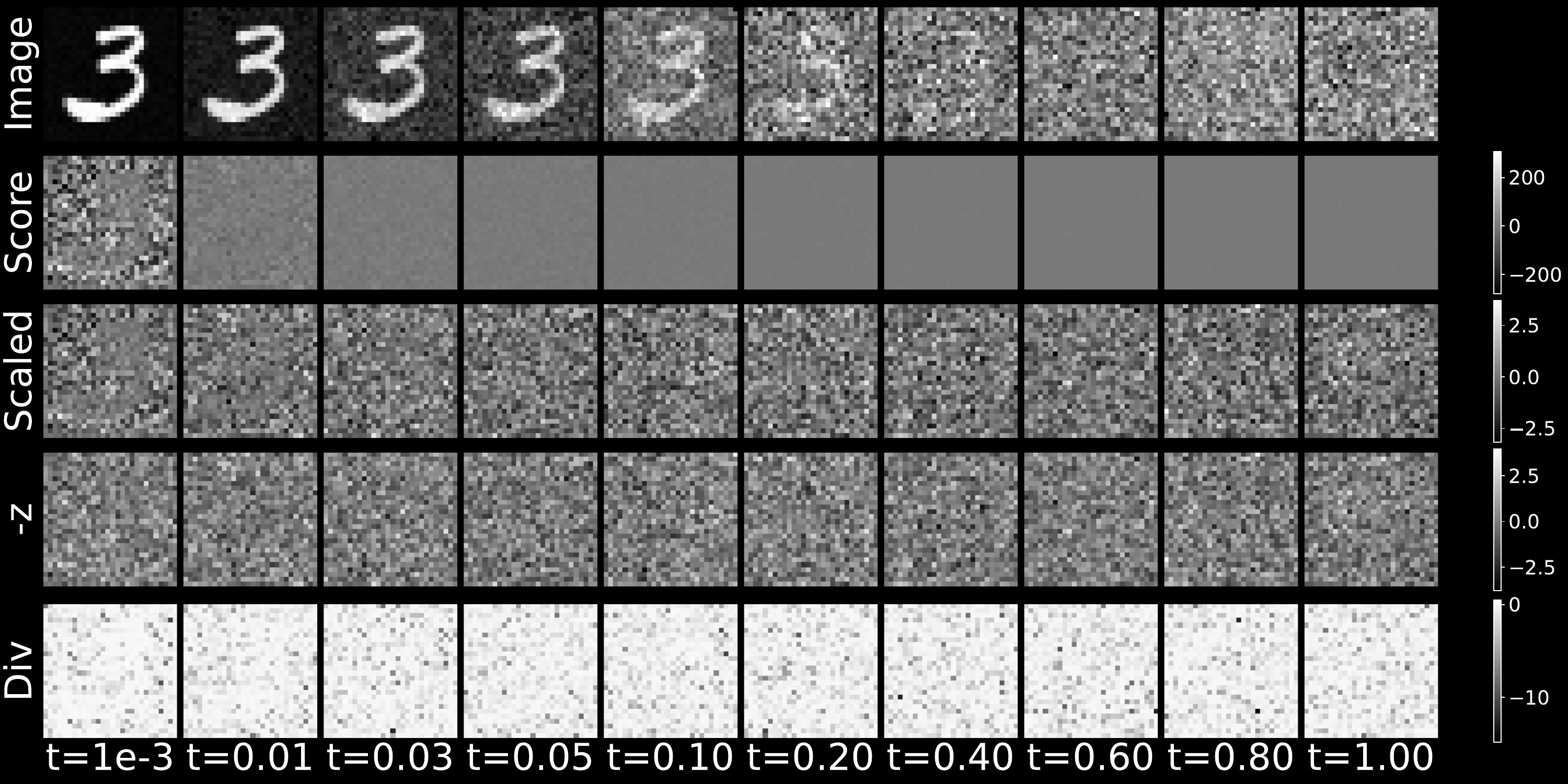}
    \caption{Noise Injection according to \eqref{FORW} in the Forward Process (1st row). The learned scores are much larger for lower noise levels (2nd row), but multiplying by $\sigma(t)$ maps them to the same scale (3rd row) and we see that they resemble the noise $\bm{z}$ (4th row). The pixel-wise contribution to the divergence multiplied by $\sigma(t)^2$ is negative (5th row).}
    \label{fig:single_score_div}
\end{figure}

To get a better understanding of the FP equation and the terms it involves, we will analyze in more detail the score and the divergence of the score and their role in the training dynamics of DMs. The divergence is the trace of the Jacobian of the score
\begin{equation}\label{div}
    \mathrm{div}_{\bm{x}} \, \bm{s}(\bm{x},t;\bm{\theta})=\sum_{i=1}^D\frac{\partial s_i(\bm{x},t;\bm{\theta})}{\partial x_i} 
\end{equation}
and we know its value, if the model approximates the score very well (i.e. $L_{\text{DSM}}(\bm{\theta}) \approx 0$). In such a situation the score model has learned to output the noise, so for a realization of $z \sim \mathcal{N}(0, I)$ at noise level t, the model would output $\bm{s}(\bm{x}(t), t; \bm{\theta})\approx-\bm{z}/\sigma(t)$ and the divergence of the model would be $\mathrm{div}_{\bm{x}} \, \bm{s}(\bm{x},t;\bm{\theta})=\sum_{i=1}^D \partial s_i(\bm{x},t;\bm{\theta})/\partial x_i\approx-D/\sigma(t)^2$, since $\bm{z}=(\bm{x}(t)-\bm{\mu}(t))/\sigma(t)$. 
This is verified empirically in Fig.~\ref{fig:single_score_div} where the scores over noise levels calculated by a noise conditional score model trained on \eqref{LDSM} in the VP framework are plotted. 
The third row shows $\sigma(t)\,\bm{s}(\bm{x}(t), t; \bm{\theta})$ and we see that it resembles $-z$ in the fourth row, meaning the model learned to predict the noise, as expected after a converged training. 
The bottom row shows the pixel-wise contributions to the divergence, mapped to the same scale: $\sigma(t)^2\, \partial s_i(\bm{x},t;\bm{\theta})/\partial{x_i}$ and we see that the majority is negative and more or less close to the value of $-1$, except some outliers.

\begin{figure}[tb]
    \centering
    \includegraphics[width=\linewidth]{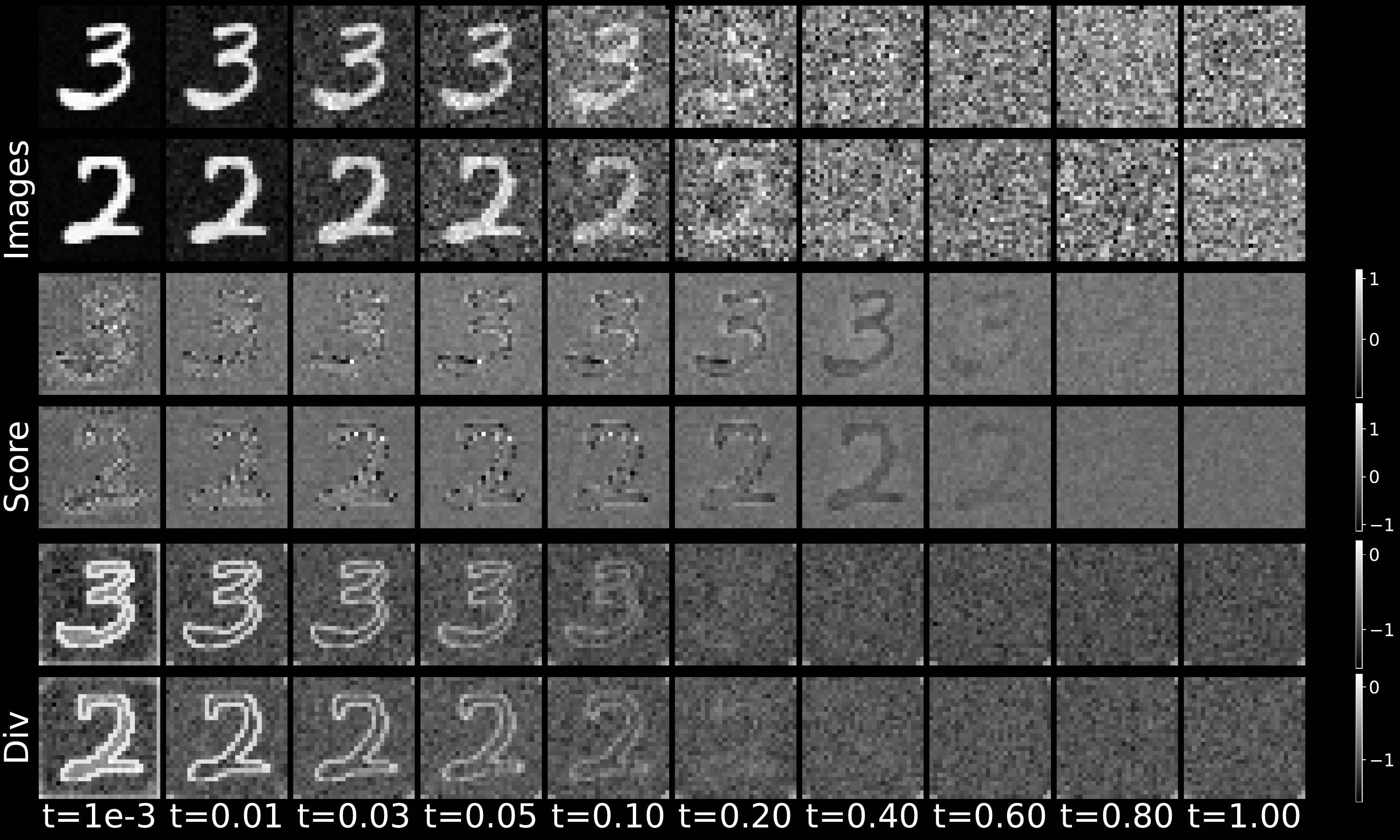}
    \caption{The first two rows show noise injection according to \eqref{FORW}. Rows 3 \& 4, and 5 \& 6 are the average of the score and the divergence, respectively, calculated over 200 realizations of the noise $Z$.}
    \label{fig:avg_score_div}
\end{figure}

To understand how scores as noisy as the ones in Fig.~\ref{fig:single_score_div} still point towards realistic images, we can take the average over several samples from $Z$ and realize that there is some signal in the score and divergence. 
This is shown in Fig.~\ref{fig:avg_score_div}.
The score plots in rows 3 and 4 indicate that at higher noise levels around $t=0.4$ the score is homogeneous and larger scores are assigned to the immediate surroundings of the digits than to the digit pixels themselves. At lower noise levels the score gets more granular and sometimes differs greatly between neighboring pixels. This is in line with the accepted understanding that in generation, DMs focus on the coarse structure first and then on the details \cite{park2023coarse}. At early stages of the generation, the model indicates that a homogeneous intensity increase of the pixels in the middle leads to higher likelihood. Since the digit pixels are already of higher intensity, they appear darker. At lower noise levels the model focuses on fine details and we see accentuated individual pixels rather than patches.

Visualizing the divergence can provide insights into how DMs learn to denoise at different stages of the process. The bottom rows of Fig.~\ref{fig:avg_score_div} show the pixel-wise contributions to the global divergence \eqref{div}, $\partial s_i(\bm{x},t;\bm{\theta})/\partial x_i$. The interpretation of these pixel-wise divergences is directly related to what the model considers likely/realistic changes in intensity: if we were to increase the intensity of one pixel, keeping all others the same, how would that change the score? In other words, a divergence of $-1$ indicates that the score would counteract the intensity increase, while a divergence of 0 means the score of pixel $i$ does not react to an increase in intensity of that pixel. 

Fig. \ref{fig:avg_score_div} shows that at high noise levels, when there is little information in the images, the divergence contains almost no information. Around $t=0.4$, we see some lighter patches in the middle of the frame, indicating that the score model knows that higher pixel intensities are more typical in the center than on the edges of MNIST images. At $t=0.2$ and $t=0.15$ we start to see the first vague outlines of the digits. In the next frames, the edges of the figures light up. 
The fact that the edges light up and the interior does not, indicates that the score model is less confident about intensities on the edges, while knowing that pixel neighborhoods on the interior are more similar and thus counteracting an increase in intensity more strongly. 

\subsection{Penalizing the squared norm of the score}
The first proposed penalty is the squared norm of the score
\begin{equation}\label{PSN}
    P_{SN} = \frac{1}{D}\,\mathbb{E}_{\bm{x}(0),t,\bm{z}} \left\| \bm{s}(\bm{x},t; \bm{\theta})\right\|_2^2.
\end{equation}
An important advantage of this penalty is that the computational cost of training the model is basically the same as the vanilla DSM loss, as the norm of the score is already computed in~\eqref{LDSM}. 
A second advantage is that this term has a clear interpretation: it simply penalizes large score values in any pixel. As shown in Fig.~\ref{fig:single_score_div}, row 2, the magnitude of the score is significantly larger at small noise levels. As this is the part of the process where also the largest deviations from the FP equation have been observed, we hope to reduce the FP residual by penalizing extreme score approximations at low $t$. 

\subsection{Penalizing the squared Frobenius norm of the Jacobian of the score}
Another proposed penalty with a similar motivation is the Frobenius norm of the Jacobian of the score. The Jacobian is the matrix of all pixel-wise partial derivatives and describes the effects of any change in the input images on the score model output. The Frobenius norm of this matrix is used as a measure of network complexity in \cite{dockhorn2022score} and is observed to increase drastically at low noise levels.
\begin{equation}\label{PJAC}
    P_{JAC} = \frac{1}{D}\,\mathbb{E}_{\bm{x}(0),t,\bm{z}} \left\| \nabla_{\bm{x(t)}} \bm{s}(\bm{x}(t),t;\bm{\theta}) \right\|_F^2
\end{equation}
As high score model complexity and high FP residuals coincide at small noise levels, we hope to improve the FP residual by explicitly penalizing model complexity.

\subsection{Penalizing the divergence of the score}
The final proposed penalty is the divergence of the score 
\begin{equation}
    P_{DIV} = \frac{1}{D}\,\mathbb{E}_{\bm{x}(0),t,\bm{z}} \left( \mathrm{div}_{\bm{x}} \, \bm{s}(\bm{x},t;\bm{\theta})\right)^2.
\end{equation}
The divergence is the trace of the Jacobian and we expect a similar effect to the above penalty \eqref{PJAC}, but at lower computational cost. On top of that, the bottom row in Fig.~\ref{fig:single_score_div} clearly shows great pixel-wise differences in the divergence. While most pixels are bright and only slightly negative, some have values below $-10$ and thus differ significantly from the value of $-1$, expected for exact score matching (see section on understanding div for motivation).

\subsection{Efficient calculation of Jacobian and divergence} 
To efficiently compute divergences in $P_{FP}$ and $P_{DIV}$ and to compute the Jacobian of the score, we employ Hutchinson's stochastic trace estimator \cite{hutchinson1989stochastic}. The divergence is the trace of the Jacobian and can be approximated as follows
\begin{equation}
    \mathrm{div}_{\bm{x}} \, \bm{s}(\bm{x},t) \approx \frac{1}{K} \sum_{k=1}^{K} v_k \bigl(\nabla_{\bm{x}} \bm{s}(\bm{x},t)\bigr) v_k^{\top},
\end{equation}
where $v_k \sim \mathcal{N}(\bm{0},\bm{I})$.
Similarly, the whole Jacobian can be estimated by
\begin{equation}
    \bigl\|\nabla_{\bm{x}} \bm{s}(\bm{x},t) \bigr\|_F^2 \approx
    \frac{1}{K} \sum_{k=1}^{K} \left\|  v_k ^{\top} \bigl(\nabla_{\bm{x}} \bm{s}(\bm{x},t)\bigr)
    \right\|_2^2,
\end{equation}
where $v_k \sim \mathcal{N}(\bm{0},\bm{I})$.
In both approximations we set $K=1$, following \cite{Lai2023FPDiffusion} and \cite{song2021sde}.

\section{Experiments}
In order to analyze the proposed penalties, multiple experiments were carried out and the results compared to the baseline DSM model.
All experiments are conducted on the MNIST dataset, which is composed of 70{,}000 grayscale images of 28$\times$28 pixels, of which 60{,}000 images were used for training the models.

The neural networks used in the experiments follow a U-Net architecture with the same configuration as the one defined in \cite{song2021sde}. 
The networks are trained using the DSM loss in \eqref{LDSM} plus a weighted penalty term
\begin{equation}
L = L_{\text{DSM}} + \lambda \, P,
\end{equation}
where $P$ denotes each one of the proposed penalty terms and $\lambda$ controls the strength of the regularization. The baseline model is trained using \eqref{LDSM} without any penalty.

The models are trained in the variance-preserving SDE framework for 200 epochs with a learning rate of $5 \times 10^{-4}$, except for the FP penalty, for which the learning rate is $1 \times 10^{-3}$, following \cite{Lai2023FPDiffusion}.

To evaluate the quality of the generated images, the following four metrics are used: (1) the Fréchet Inception Distance (FID) as a measure of overall perceptual quality; (2) Density, which measures how close the generated images are to the real ones in the space of extracted features \cite{density_coverage2020}; (3) Coverage, which measures the mode coverage of the generated images with respect to the training set \cite{density_coverage2020}; and (4) the Shannon entropy of the class distribution as an additional measure of diversity. For every trained model, 10{,}000 samples are generated and compared with the MNIST test set. 
The computation of the FID score requires extracting and comparing features from real and generated images. As MNIST differs from the training data of the Inception network, we use a pre-trained LeNet classifier for feature extraction.
The Shannon entropy is computed on the class distribution given by the LeNet classifier over the generated images. 

To evaluate these metrics, 20 training runs were performed for all models.
The average performance on these four metrics and the training time per epoch is reported in Table~\ref{tab:overview}. Training was conducted on a machine with an NVIDIA GeForce RTX 5070 GPU and an AMD Ryzen AI 9 365 CPU, using Pytorch 2.9.  

\begin{table*}[tb]
    \centering
    \caption{Comparison of penalty terms on MNIST. $P_{SN}$ and $P_{DIV}$ improve performance with minimal computational overhead.}
    \label{tab:overview}
    \begin{tabular}{cccccc}
        \toprule
        \textbf{Loss} 
        & \textbf{FID (↓)} 
        & \textbf{Density (↑)} 
        & \textbf{Coverage (↑)} 
        & \textbf{Entropy (↑)}
        & \textbf{Time/epoch (s) (↓)} \\
        \hline
        $L_{DSM}$ 
        & 20.32 $\pm$ 11.41
        & 80.10 $\pm$ 5.20 
        & 85.69 $\pm$ 1.89 
        & 2.263 $\pm$ 0.023  
        & 6.25 \\
        $+P_{FP}$ 
        & 16.25 $\pm$ 7.16 
        & 82.85 $\pm$ 3.71 
        & 87.97 $\pm$ 1.67  
        & 2.269 $\pm$ 0.017  
        & 12.32 \\
        $+P_{SN}$ 
        & 17.50 $\pm$ 7.89 
        & 80.44 $\pm$ 4.28 
        & 87.11 $\pm$ 1.59  
        & 2.265 $\pm$ 0.022
        & 6.36 \\
        $+P_{JAC}$ 
        & 17.02 $\pm$ 7.51 
        & 81.64 $\pm$ 4.94 
        & 87.29 $\pm$ 1.75
        & 2.268 $\pm$ 0.0189
        & 19.86 \\
        $+P_{DIV}$ 
        & 18.46 $\pm$ 8.88 
        & 80.25 $\pm$ 4.57 
        & 86.61 $\pm$ 1.74
        & 2.264 $\pm$ 0.023
        & 7.34 \\
        \bottomrule
    \end{tabular}
\end{table*}
All tested regularizers improve the baseline performance on average in all four metrics. The FP penalty performs best with an average FID of 16.25 compared to 20.32 without regularization, but also the Jacobian and the squared norm penalty get close to this performance, with 17.02 and 17.50 on average, respectively. In terms of training times and computational expenses, the squared norm stands out. With 6.36 s/epoch it takes only marginally longer than the baseline with 6.25 s/epoch. The FP penalty, on the other hand, takes about twice as long at 12.32 s/epoch. This difference is expected to be more severe with more complex datasets, as was found by \cite{Lai2023FPDiffusion}.

\begin{figure}[tb]
    \centering
    \includegraphics[width=\linewidth]{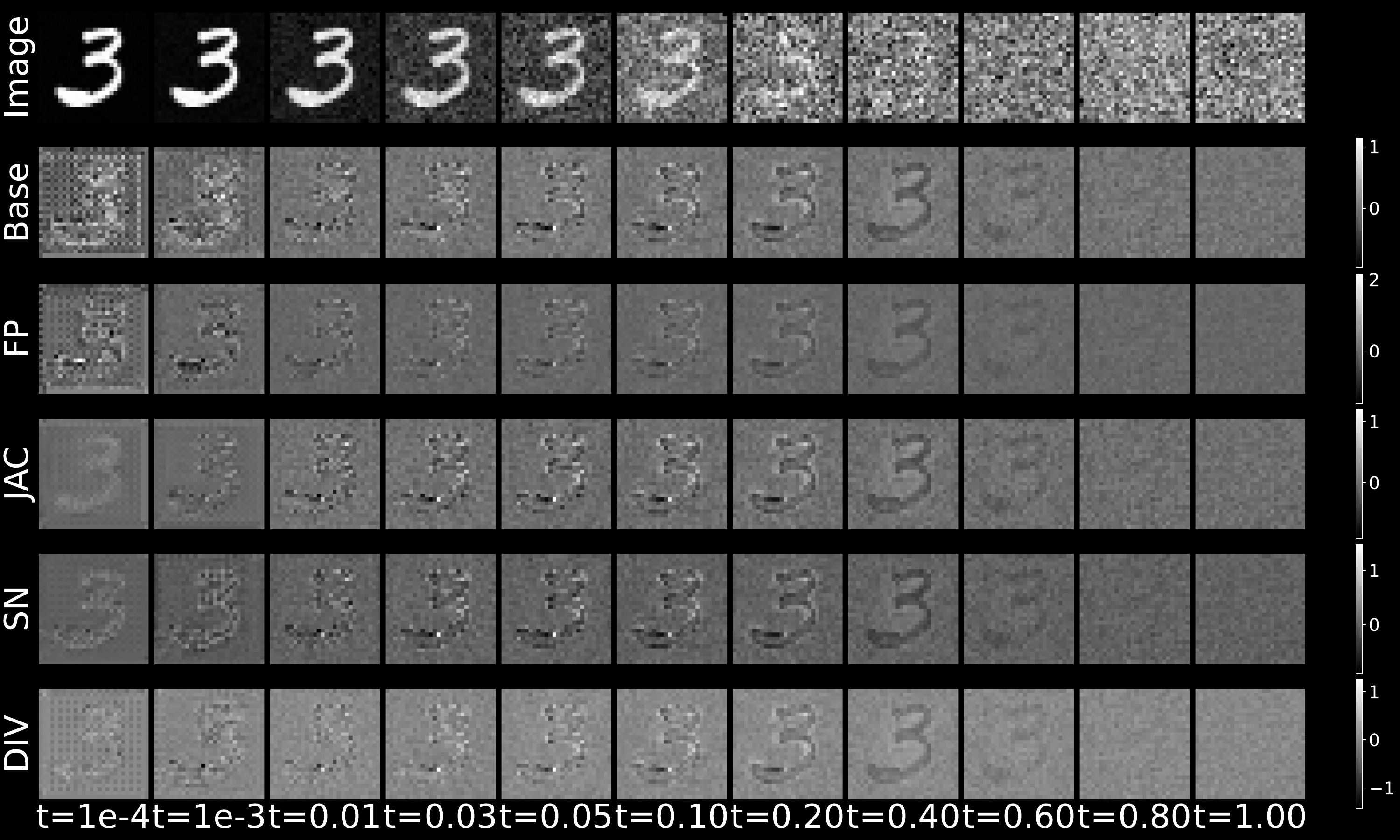}
    \caption{Score depending on penalty term}
    \label{fig:s_comp}
\end{figure}

\begin{figure}[tb]
    \centering
    \includegraphics[width=\linewidth]{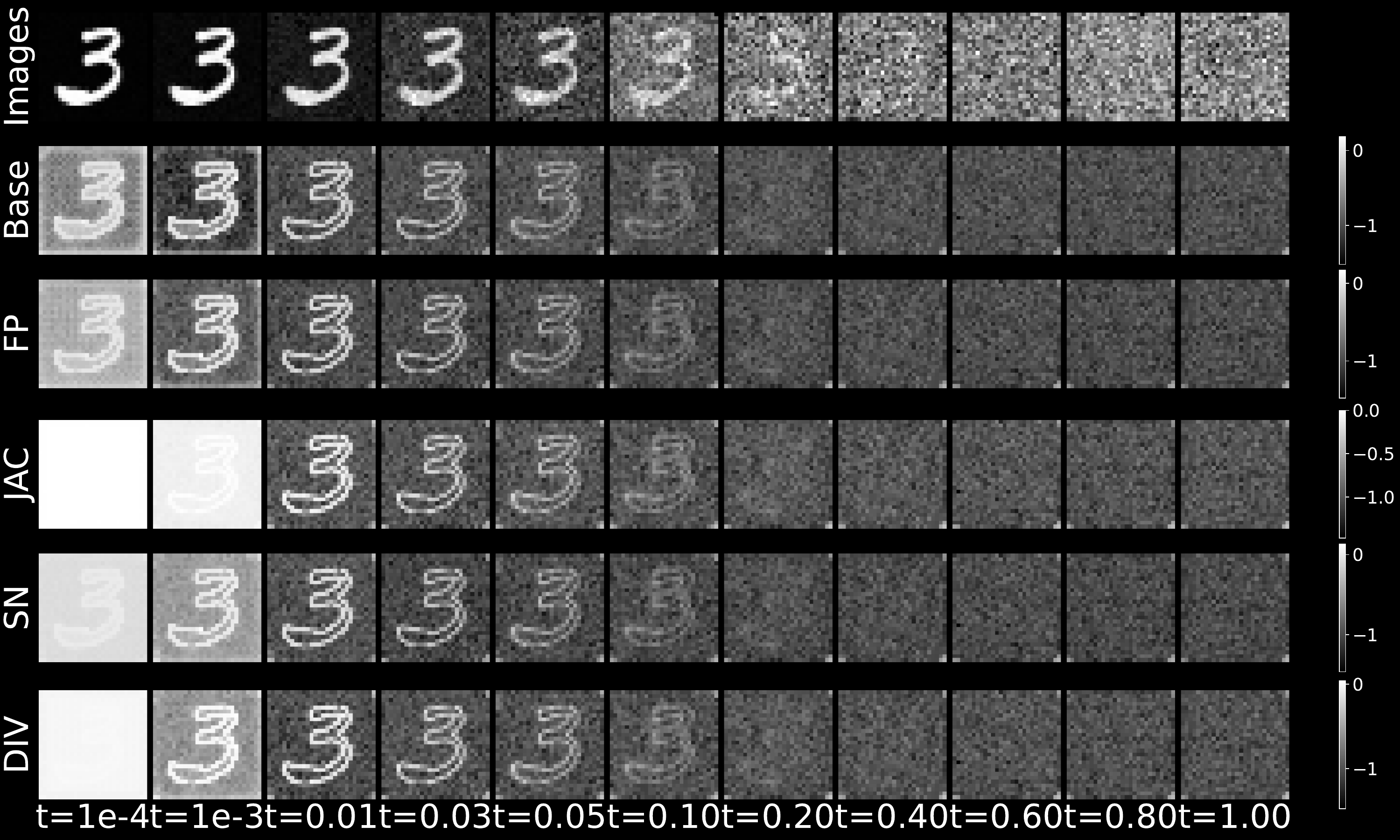}
    \caption{Divergence depending on penalty term}
    \label{fig:div_comp}
\end{figure}

To more closely examine the behavior of models trained with different penalties, the score and the divergence at different stages of the noising process are plotted in Figs.~\ref{fig:s_comp} and \ref{fig:div_comp}, respectively. Although all models show similar behavior at high noise levels, there are striking differences at low ones. All regularization terms seem to suppress the attention to detail that the score maps of the baseline show at $t=1e-3$ and $t=1e-4$. Instead, the regularized score maps look more homogeneous, similar to higher noise levels. 

In the divergence evolution in Fig.~\ref{fig:div_comp} these differences are even more extreme. The Jacobian and the divergence penalty seem to make the model completely indifferent to intensity changes of individual pixels.

\begin{figure}[tb]
    \centering
    \includegraphics[width=\linewidth]{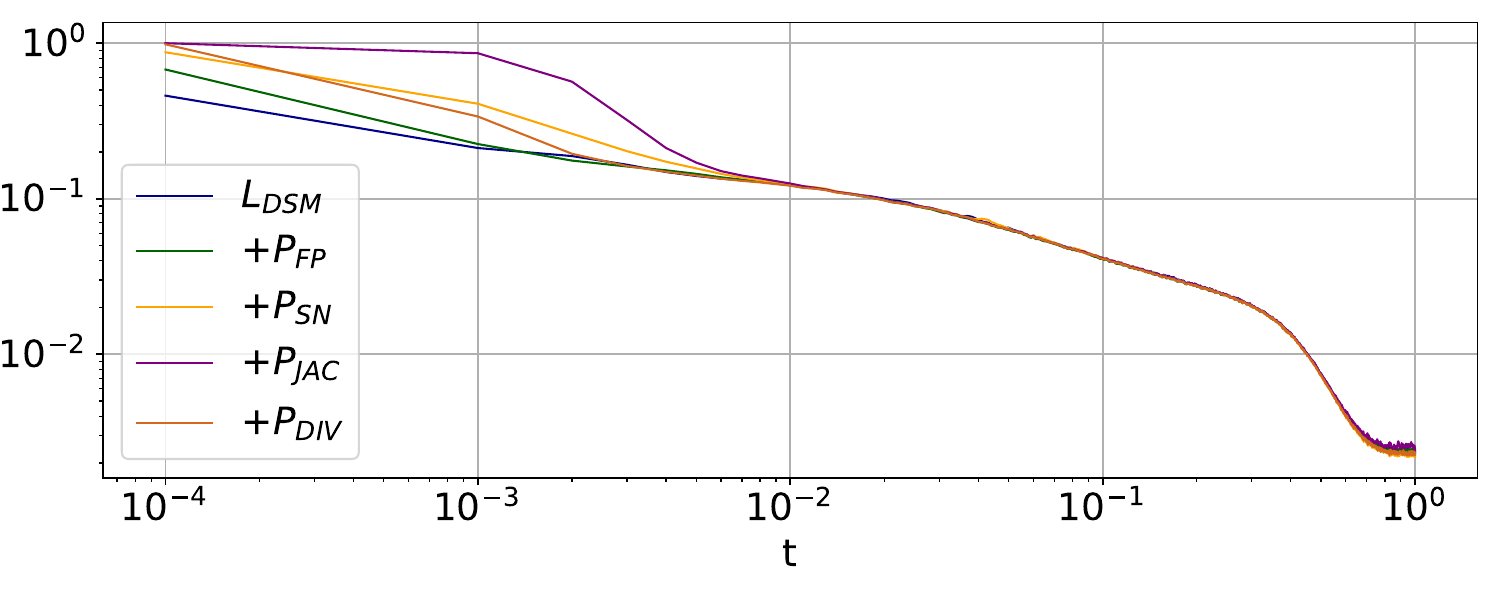}
    \caption{The DSM loss over noise levels, evaluated for models trained with different penalty terms.}
    \label{fig:dsm_pens}
\end{figure}

\begin{figure}[tb]
    \centering
    \includegraphics[width=\linewidth]{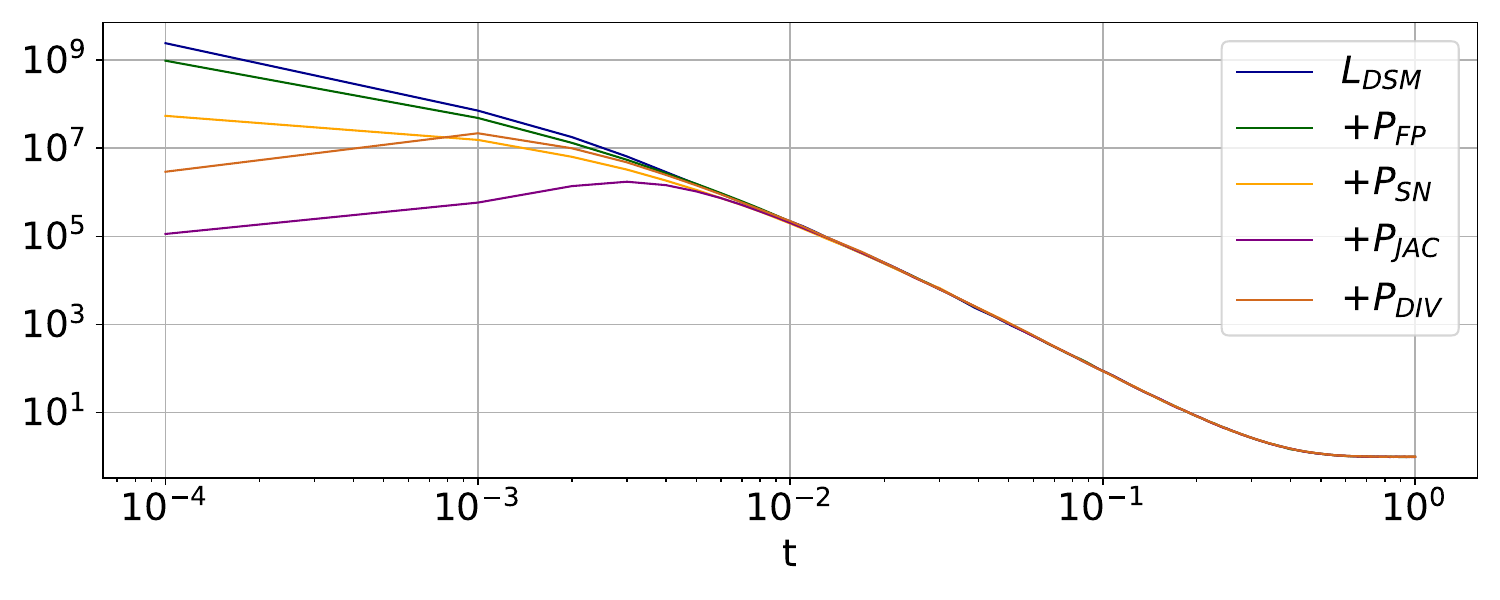}
    \caption{The Frobenius norm of the Jacobian, evaluated for models trained with different penalty terms.}
    \label{fig:jac_pens}
\end{figure}

The strong effect of regularization at small noise levels can also be observed in the plots of the DSM loss (Fig.\ref{fig:dsm_pens}) and the Frobenius norm of the Jacobian (Fig.~\ref{fig:jac_pens}). As reported in \cite{dockhorn2022score} and \cite{debortoli2024targetscorematching}, both increase exponentially at low noise levels without regularization. Compared to the baseline all penalties accelerate the increase in $L_{DSM}$ at low noise, while reducing the Frobenius norm of the Jacobian. The strongest effect in both plots is observed for the penalty on the Frobenius norm of the Jacobian itself. 

\begin{figure}[tb]
    \centering
    \includegraphics[width=\linewidth]{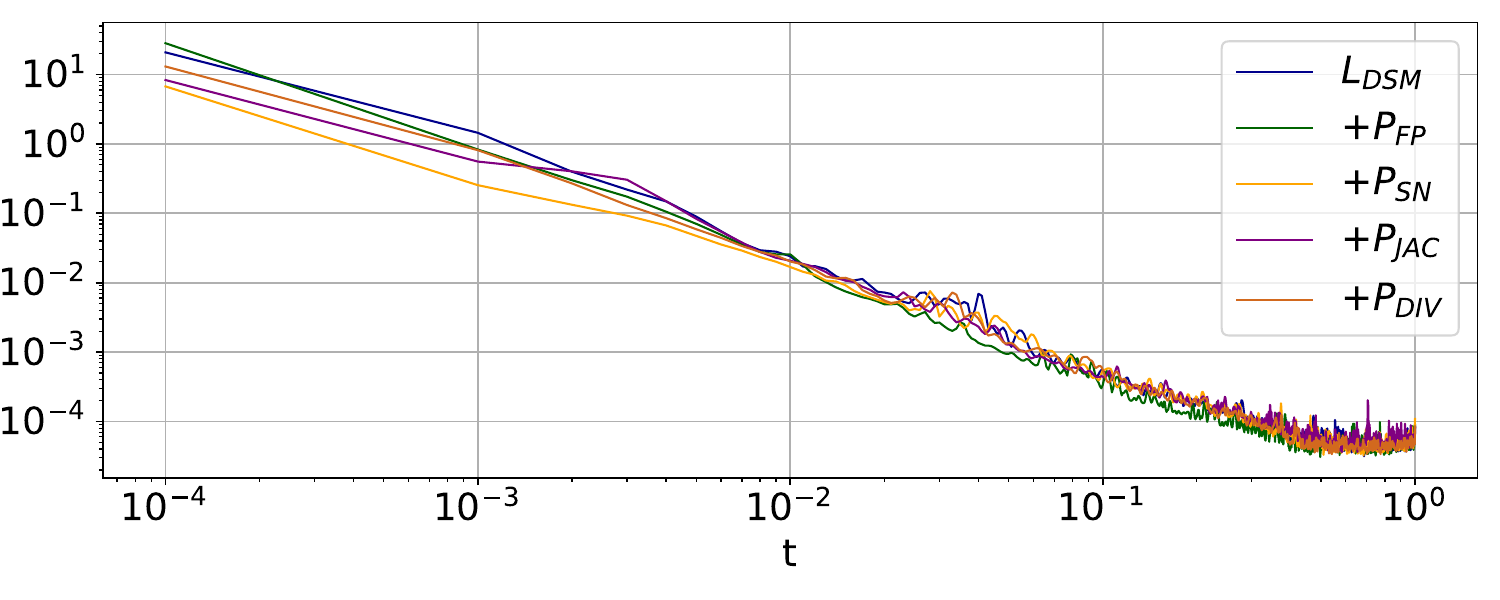}
    \caption{The FP residual over noise levels, evaluated for models trained with different penalty terms.}
    \label{fig:fp_pens}
\end{figure}

\begin{figure}[tb]
    \centering
    \includegraphics[width=\linewidth]{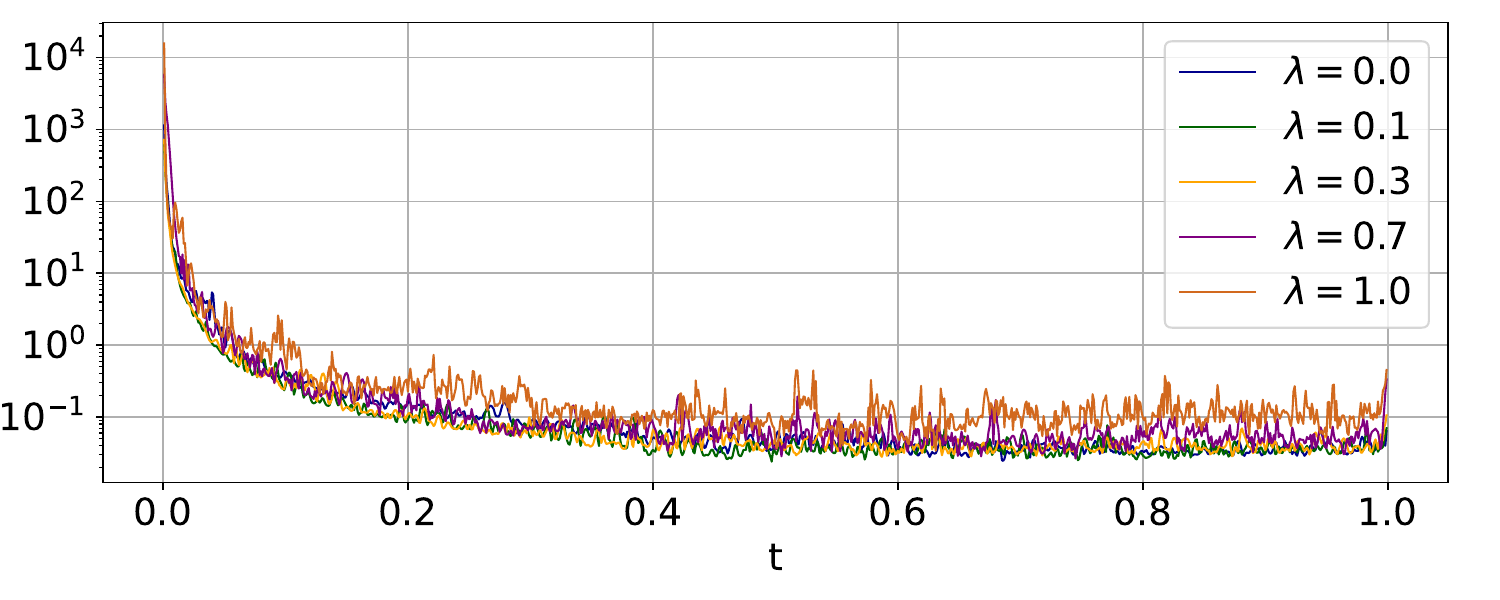}
    \caption{FP Residual reduction depending on regularization strength}
    \label{fig:abla}
\end{figure}

With regards to the FP residual we observe that it is difficult to achieve a strong reduction. Fig.~\ref{fig:fp_pens} indicates that all penalty terms reduce the FP residual, especially at small noise levels, but all types of regularization do not make a big difference. In Fig.~\ref{fig:abla} we see that aiming for a closer adherence to the FP equation by setting a higher $\lambda$ results in an even worse FP residual. This counterintuitive observation strengthens the hypothesis that the benefits of FP regularization stem from general regularization effects rather than from specifically enforcing the FP equation.


\section{Conclusion}

We studied the role of regularization in DMs from the perspective of the Fokker--Planck equation. Enforcing the score FP equation via an explicit penalty reduces FP residuals and improves the quality of the generated images, but so do all other tested penalties. The counterintuitive finding that stronger FP regularization produces worse FP residuals than moderate regularization is consistent with the idea that the benefits of FP regularization stem from general regularization effects rather than specifically enforcing the FP equation. Our experiments show that regularization most strongly affects small noise levels, which is sensible, as that is where the penalized quantities, network capacity, score magnitude and FP deviation, are largest. 

We conclude that simple penalty terms can have similar benefits to the complex FP penalty, one of them at negligible computational cost. Our analysis is limited to MNIST and validating the observations on other datasets will be important. We expect that this will strengthen our claims, as MNIST is relatively simple and the computational overhead of regularization is expected to increase with dataset complexity.

\bibliographystyle{IEEEtran} 
\bibliography{references}

\end{document}